\documentclass[conference]{IEEEtran}
\IEEEoverridecommandlockouts
\usepackage{cite}
\usepackage{amsmath,amssymb,amsfonts}
\usepackage{algorithmic}
\usepackage{graphicx}
\usepackage{textcomp}
\usepackage{float}
\usepackage {subcaption}
\usepackage{xcolor}
\usepackage[section]{placeins}
\def\BibTeX{{\rm B\kern-.05em{\sc i\kern-.025em b}\kern-.08em
		T\kern-.1667em\lower.7ex\hbox{E}\kern-.125emX}}
\begin{document}
	
	\title{Local Adaptive Clustering Based Image Matching for Automatic Visual Identification}
	
	\author{
		
		\IEEEauthorblockN{Zhizhen Wang}
		\IEEEauthorblockA{\textit{College of Information Science and Technology},\\
			Beijing University of Chemical Technology,\\
			Beijing, China.\\
			2023200830@mail.buct.edu.cn}\\}
	\maketitle
	
	\begin{abstract}
		Monitoring cameras are extensively utilized in industrial production to monitor equipment running. With advancements in computer vision, device recognition using image features is viable. This paper presents a vision-assisted identification system that implements real-time automatic equipment labeling through image matching in surveillance videos. The system deploys the ORB algorithm to extract image features and the GMS algorithm to remove incorrect matching points. According to the principles of clustering and template locality, a method known as Local Adaptive Clustering (LAC) has been established to enhance label positioning. This method segments matching templates using the cluster center, which improves the efficiency and stability of labels. The experimental results demonstrate that LAC effectively curtails the label drift.
	\end{abstract}
	
	
	\section{Introduction}
	With the development of computer vision and the requirements of industrial processing, video surveillance and real-time analysis of the screen have been widely used in industrial inspection. Many industries have proposed and applied digital workshop, using Internet of things technology to collect data which real-time production produces, optimize production process, and achieve agile response and rapid recovery of abnormal events in the production process\cite{b1}.

	In industrial production, we usually mark the equipment so that the information of the corresponding equipment can be displayed in the camera picture. The most important task is how to map the devices in the camera image to them in the panorama. In the method based on image feature point detection and matching\cite{b2}, the core idea is to obtain the image template of the workshop equipment in advance, and then detect and match the feature points between the template image and the real-time image, so as to determine the mapping relationship between two images.

	Image feature point extraction is a basic task in the field of computer vision. The classical SIFT (Scale-invariant feature transform)\cite{b3} and SURF (Speeded Up Robust Features)\cite{b4} algorithms use Gaussian pyramid to construct scale space, and each layer of the pyramid will lead to different degrees of boundary information loss, which affects the accuracy of recognizing. ORB (Oriented FAST and Rotated BRIEF)\cite{b5} algorithm combines FAST and BRIEF\cite{b6} to improve the calculation speed, but it lacks scale invariance. KAZE\cite{b7} algorithm uses AOS numerical approximation method to establish nonlinear scale space, which has a large amount of calculation and is difficult to meet the real-time performance.

	For feature point matching, RANSAC (Random Sample Consensus)\cite{b8} method is mainly used in the early stage, and the correct model parameters are selected from the data set by continuous iteration. However, the method is time-consuming and the accuracy is difficult to guarantee. The GMS (Grid-based Motion Statistics)\cite{b9} algorithm uses the principle of motion consistency to improve the matching accuracy, and uses the grid framework to improve the speed. SuperPoint\cite{b10} and SuperGlue\cite{b11} in deep learning have achieved significant improvement in accuracy, but they rely on high computing power support. Therefore, this paper adopts the traditional image feature point extraction and matching algorithm, and uses the spatial locality principle to improve the processing speed and reduce the label drift. Its advantage is that it can run fast without high performance equipment, and the results are interpretable during the algorithm. This allows us to evaluate the algorithm performance on a process-by-process basis to achieve local performance optimality and ultimately improve the overall efficiency.

	Real-time Homography Mapping\cite{b12} is an important technology in computer vision, and it is also the key to image matching. It can map an image from one reference coordinate system to another target coordinate system. It is widely used in mobile cameras, such as autonomous vehicles, robot navigation and augmented reality.

	The theoretical basis of real-time homography mapping technology is the homography matrix\cite{b13}. This matrix maps the coordinates of a point in one coordinate system to the coordinates of another coordinate system. In real-time homography mapping, we need to calculate the homography matrix between the current image and the target image, and then use the matrix to transform the current image, and finally get the current image in the coordinate system of the target image. In practical applications, real-time homography mapping technology still needs to solve some problems, such as noise interference, motion blur and so on. Therefore, some additional algorithms should be used to optimize the performance and stability of real-time homography mapping techniques\cite{b14}.

	In the actual homography mapping, we first need to extract the feature points of the two images, and then match the feature points of the two images to obtain the paired feature point pairs. Using some outlier filtering methods, we remove a part of the outliers from the feature point pairs and estimate the homography matrix of the transformation from them. Through the analysis of the actual inspection surveillance video, we found that the camera was continuously changing and there was no sudden jump in the picture during movements. In addition, the camera will stay longer in areas with more equipment\cite{b15}.

	Therefore, this paper proposes a Local Adaptive Clustering method to optimize the image matching. The matching template is segmented with the clustering point as the center\cite{b16}, which improves the efficiency and stability of image matching. The process of real-time homography mapping based on image feature matching. Firstly, in order to improve the locality of device template, we cluster device information coordinates. After that, BEBLID (Boosted Efficient Binary Local Image Descriptor) descriptor\cite{b17} and GMS algorithm are used for feature matching and outlier filtering\cite{b18}, and DEGENSAC (Degenerate Sample Consensus)\cite{b19} is used to obtain the homography matrix. Then the local matching is performed by counting the template index corresponding to the first three frames of the current frame, and the best template is selected by soft voting method\cite{b20}. Finally, by recording the relative information between labels, the coordinates are optimized by polar coordinates.

	In summary, our main contributions include:
	\begin{itemize}
		\item An efficient image matching workflow.Which are extracted by BEBLID descriptor, the feature points of image are coarse-filtered by GMS algorithm, and then fine-filtered by DEGENSAC to obtain a homography matrix with higher interior point rate.
		\item A local adaptive clustering algorithm.According to the principle of clustering locality and template locality, a template segmentation and filtering algorithm centered on cluster points is designed to optimize the matching rate and stability.
	\end{itemize}

	\section{Methodology}
	
	\subsection{Algorithm Overview}
	
	Firstly, we manually label the equipment in the panorama and construct the topological structure by using the coordinate of the label. We save the relative position, distance and angle of labels. Then we cluster the panorama according to the coordinate distribution of labels, and segment the template according to the clustering results. We extract ORB features and BEBLID descriptors from the template in advance. Then we match the features of the current picture and the template respectively and count the feature point pairs corresponding to each template. By soft voting, the template with the largest number of feature points is selected for coordinate mapping.

	In order to reduce the number of matching templates, we design a local area to store the templates corresponding to the first two frames, which can effectively improve the matching speed. After finding the corresponding template, the outer points are roughly filtered by GMS method, and then the homography matrix is generated by DEGENSAC according to the feature points of two pictures, so as to calculate the coordinates of the labels of the corresponding template in the picture. Finally, the pre-constructed topological structure is used to optimize the generated coordinates by polar coordinates.
	\begin{figure}[H]
		\centering 
		\includegraphics[height=3.5cm,width=0.50\textwidth]{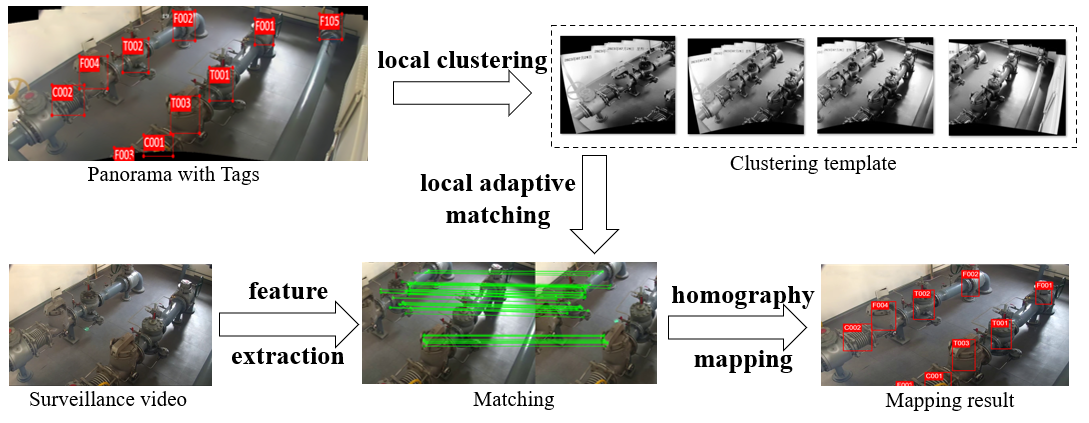}
		\caption{Flowchart of the algorithm}
		\label{2}
	\end{figure}
	
	\subsection{Principle of Locality}
	Clustering is a type of unsupervised learning\cite{b21}. It can group similar objects into the same cluster. The more similar the objects are, the better the cluster is. However, during the process of moving, the camera stays longer in the place with dense labels. In order to reduce the drift of the label after mapping, the template corresponding to the camera picture should fall in the area with dense labels as much as possible. In other words, we want each template to contain a sufficient number of labels. This method can not only ignore other unimportant feature information on the panorama, but also reduce the drift of the label caused by frequently changing the matching template.

	In this paper, K-Means algorithm is commonly used for clustering, and its basic principle is a cyclical process of moving the centre of the cluster. Specifically, we regard the coordinates of every label on the panorama as individual points and cluster them to obtain $k$ centres. Initially, a set of $k$ points is extracted from the data distribution as the initial clustering centre. Afterward, through continuous iterations, the points in each class are moved to the average position of the other points in the class. Consequently, the interior points of each class are re-divided continuously until all the points are at the closest distance from the centre of their respective class. Take an observation set $(x_1,x_2,\dots, x_n)$, which are clustered into $k$ sets $S= (S_1,S_2,\dots,S_k)$ by the K-Means algorithm to minimize the sum of squares within the group. Find the set $S_i$ that satisfies the following \eqref{eq1}:
	\begin{equation}
		argmin\sum_{i=1}^{k} \sum_{x\in S_i}^{} ||x-\mu _i||^2=argmin\sum_{i=1}^{k} |S_i|VarS_i\label{eq1}
	\end{equation}
	where $S_i$ is a cluster, $\mu_i$ denotes the centre point of $S_i$, and $x$ denotes the interior point of $S_i$.

	In order to reduce the interference of the features of other regions on the panorama to the target region, we segmented the panorama to ensure that the template image generated by segmentation has sufficient and significant features. Therefore, following assumptions are made. Firstly, when the monocular camera rotates, the image is continuous and stable, and there will be no discontinuous pictures, which is in line with the principle of spatial locality\cite{b22}. Secondly, there are more camera cruise points in the lable-dense region, and the dwell time in these regions during the rotation will be more than in the lable-sparse region.

	Based on the above assumptions, the following conclusions can be obtained. The template of the current frame is related to the template of the previous frames, and the corresponding template changes continuously because the frame changes continuously. In order to keep the template variation small, we need to make the features contained in the template as specific as possible and make the differences between the templates significant. Therefore, when we segment the template, we combine the coordinate information of the label for clustering segmentation, which can ensure that each template has enough and significant feature information.

	Then, a global local area is set to store the template index of the previous three frames, and the template of the current frame can narrow the search range according to the template in the local area, which greatly improves the matching rate. Of course, an image may correspond to two similar templates in two frames, and different templates may have a deviation effect. Therefore, we count the feature point pairs between the camera image and each template, and select the most number of point pairs through the soft voting method to ensure the uniqueness and stability of the template.
	
	\subsection{BEBLID}
	The BEBLID descriptor is a binary description algorithm used to describe the feature points. The method uses integral image to efficiently calculate the difference of average gray values in the square area of the image. The algorithm is trained using AdaBoost algorithm\cite{b23} and imbalanced dataset to address errors in retrieval and matching process. By minimizing a new similarity loss function, all weak learners share a common weight, so as to realize the descriptor binarization.

	In the algorithm, $\left(x_i,y_i,l_i\right)_{i=1}^N$ is the training set consisting of image patch pairs, $(x_i,y_i)\in X$, where $l_i\in\left\{-1,1\right\}$,$l_i$ equal to 1 indicates that two image patches correspond to the same salient image structure. $l_i$ equal to $-1$ indicates that two image patches correspond to the different salient image structure. During the training process, the loss is guaranteed to be as low as possible. The loss function is given in \eqref{eq2}:
	\begin{equation}
		\Gamma _{BEBLID}=\sum_{i=1}^{N}exp(-\gamma l_i\frac{\sum_{k=1}^{K}\alpha _kh_k(x_i)k_k(y_i) }{g_s(x_i,y_i)}) \label{eq2}
	\end{equation}
	Where $\gamma$ is the learning rate parameter and $h_k\left(z\right)\equiv h_k\left(z; f,T\right)$ corresponds to the combination of the k-th WL in the set $g_s$ and the weight $\alpha_k$, WL depends on the feature extraction function $f:X\ \rightarrow R$ and the threshold $T$. As shown in \eqref{eq3}. Given $f$ and $T$, WL is defined by thresholding $f\left(x\right)$ and $T$.
	\begin{equation}
		h(x;f,T)=\begin{cases}
			+1;f(x)\le T \\
			-1;f(x)> T
			
		\end{cases} \label{eq3}
	\end{equation}
	Where $h(x)$ is the k-th WL response vector to the image block $x$.

	The key to improve the efficiency of the BEBLID descriptor is to choose a feature extraction function that is both discriminative and fast to compute, as shown in \eqref{eq4}:
	\begin{equation}
		f(x;p_1,p_2,s)=\frac{1}{s^2}(\sum_{q\in R(p_1,s)}^{}I(q)-\sum_{r\in R(p_2,s)}^{}I(r) )\label{eq4}
	\end{equation}
	Where $I(t)$ is the gray value at pixel $t$ and $R(p,s)$ is a square image box of size $s$ centered at pixel $p$. $f$ calculate the difference between the average gray values of the pixels in $R(p_1,s)$ and $R(p_2,s)$. After $f$ is computed, the robustness of the descriptor is improved by locating and scaling the measurements with the underlying local structure. Finally, the binary descriptor is obtained by optimizing the loss function, so as to improve the real-time performance of the algorithm.
	
	\begin{figure}[H]
		\begin{minipage}[t]{0.5\linewidth}
			\centering
			\includegraphics[width=0.9\textwidth]{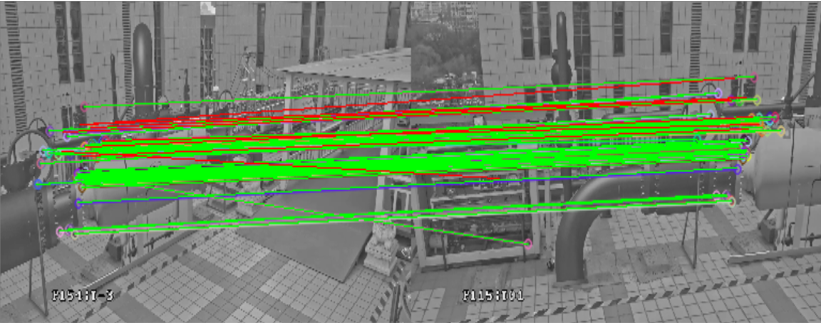}
			\centerline{(a) ORB}
		\end{minipage}%
		\begin{minipage}[t]{0.5\linewidth}
			\centering
			\includegraphics[width=0.9\textwidth]{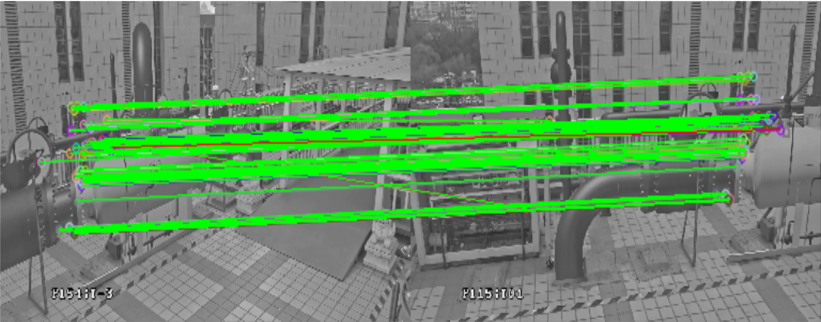}
			\centerline{(b) BEBLID}
		\end{minipage}
		\caption{(a): Feature point matching results for ORB descriptors. (b):Feature point matching results for BEBLID descriptors. The green line represents a correct match and the red line represents an incorrect match. The BEBLID descriptor outperforms the ORB descriptor in terms of matching accuracy.}
		\label{fig}
	\end{figure}
	
	\subsection{Grid-based Motion Statistics}
	The GMS algorithm is a fast robust feature matching filtering algorithm based on motion statistics. Its main idea is to divide the video frame into grids, and then perform motion statistics on the pixels in each grid. The algorithm considers that the correctly matched and incorrectly matched feature points usually do not appear in the same region. Therefore, the premise of the algorithm is that there are many other feature point pairs in the neighborhood of the correct matching, and there are very few feature point pairs in the neighborhood of the wrong matching. Therefore, we need to count the number of pairs of feature points in a certain neighborhood to determine whether it is a correct match.

	Firstly, the algorithm needs to extract a large number of feature points, and carry out a large range of feature matching to ensure that the matching samples are enough. For each matching pair, the number of matching pairs in its neighborhood is calculated to obtain the probability that the neighborhood supports the matching pair. In order to improve the speed of the algorithm, the image is divided into 20×20 grids by default, and the calculation for the matching pair is changed to the calculation for the grid. The matches that fall into the same grid are regarded as the neighborhood, and the matches that fall into two grids at the same time are regarded as the similar neighborhood. The algorithm divides the grid as small as possible, and uses the convolution kernel to include more neighborhood information, as shown in the figure. At the same time, in order to solve the problem of multi-scale kernel and multi-rotation, multiple convolution kernels with different scales and different directions are used to extract neighborhood information at the same time.
	
	\section{Experiments}
	This experiment is implemented on the PC side using the Centos7 system. In terms of hardware configuration, the CPU uses 12th Gen Intel i7-12700KF, the graphics card is NVIDIA GeForce RTX 3090, and the memory is 24000MB. In terms of software configuration, Anaconda is used for package management and C++ and Python3.7 are used for code writing. Real-time homography mapping uses PyCharm, using Python3.7 and OpenCV4.7.2 to achieve functionality. Higher versions of OpenCV update many modules with better results, such as the BEBLID descriptor and the GMS algorithm, as well as the AKAZE algorithm. The experimental video uses the industrial environment in the actual workshop to better simulate the real environment.
	
	\subsection{Interior Point Rate Comparison}
	In order to verify the efficiency of feature point extraction and matching, we design four sets of comparative experiments to illustrate the advantages of BEBLID, GMS and DEGENSAC algorithms. Firstly, 2500 feature points are extracted using ORB, and then feature matching is performed with the descriptor of ORB and the descriptor of BEBLID, and the homography matrix is calculated using RANSAC,GMS+ RANSAC,GMS+DEGENSAC respectively. Through the returned mask matrix M, the interior point rates of both are obtained and the running time is counted. The experiment was performed ten times and the results were averaged as shown in Table 1.

	\begin{table}[htbp]
		\caption{Comparison of interior point rate and computation time}
		\begin{center}
			\begin{tabular}{|c|c|c|}
				\hline
				\textbf{Algorithm Combination}&{\textbf{Interior Point Rate}} &{\textbf{Time}}\\
				
				\hline
				ORB+RANSAC& $16.18\%$ &0.16s\\
				\hline
				ORB+GMS+RANSAC& $66.86\%$ &0.22s\\
				\hline
				ORB+BEBLID+GMS+RANSAC& $69.77\%$ &0.33s\\
				\hline
				ORB+BEBLID+GMS+DEGENSAC& $92.03\%$ &0.39s\\
				\hline
			\end{tabular}
			\label{tab1}
		\end{center}
	\end{table}
	We can see that in the experimental comparison between the first group and the second group, filtering the outer points by adding GMS can greatly improve the inner point rate. Through the experimental comparison between the second and third groups, we can obtain that the BEBLID descriptor can improve the inlier point rate to some extent. Through the experimental comparison of the third and fourth groups, we get that the DEGENSAC method is better than the RANSAC method, and the generated homography matrix is more in line with the actual situation.
	\subsection{Cluster Drift Comparison}
	The drift is reflected by recording the change of the relative position coordinate of the label in panorama. Different colors represent different labels. The ordinate is the relative abscissa of each label, and the abscissa is the time. The monitoring footage of the actual workshop was used for the test video. We can see that the drift situation is improved after modification. To make the difference before and after improvement more obvious when calculating the variance, we scale up the relative coordinates by a factor of 100.
	
	\begin{figure}[htbp]
		\begin{minipage}[t]{0.5\linewidth}
			\centering
			\includegraphics[width=\textwidth]{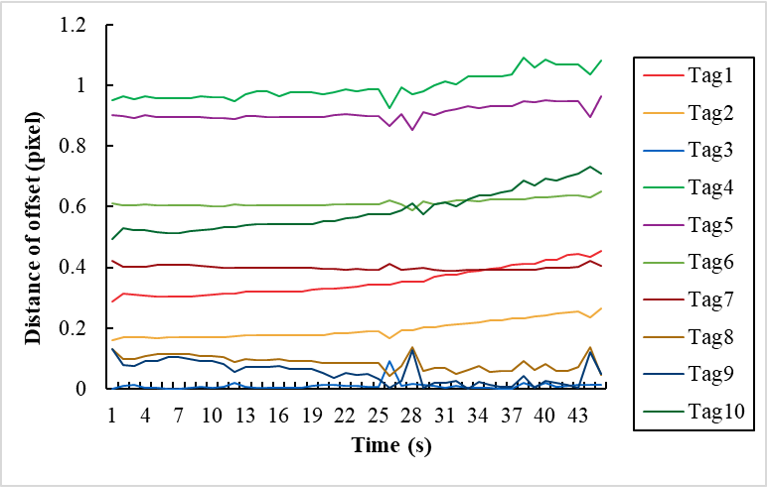}
			\centerline{(a) before}
		\end{minipage}%
		\begin{minipage}[t]{0.5\linewidth}
			\centering
			\includegraphics[width=\textwidth]{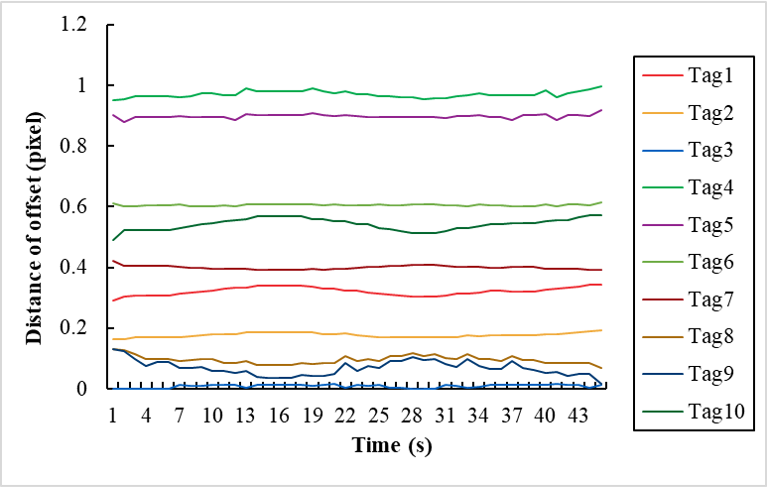}
			\centerline{(b) after}
		\end{minipage}
		\caption{Comparison of clustering effect}
		\label{fig}
	\end{figure}
	We can see that the improved abscissa moves more smoothly. Among them, label 3 and label 4 are significantly improved. Label 7, label 8 is improved to some extent. Then, it can be seen from the variance that the overall variance decreases, indicating that the overall fluctuation situation has been optimized to some extent.

	Besides we calculate the Euclidean distance of each coordinate in the current frame and the previous frame, which can better reflect the movement of the same coordinate in two frames.
	\begin{figure}[htbp]
		\begin{minipage}[t]{0.5\linewidth}
			\centering
			\includegraphics[width=\textwidth]{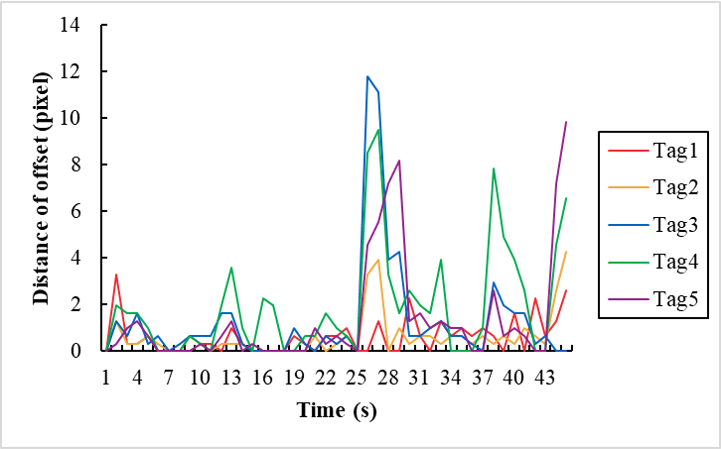}
			\centerline{(a) before}
		\end{minipage}%
		\begin{minipage}[t]{0.5\linewidth}
			\centering
			\includegraphics[width=\textwidth]{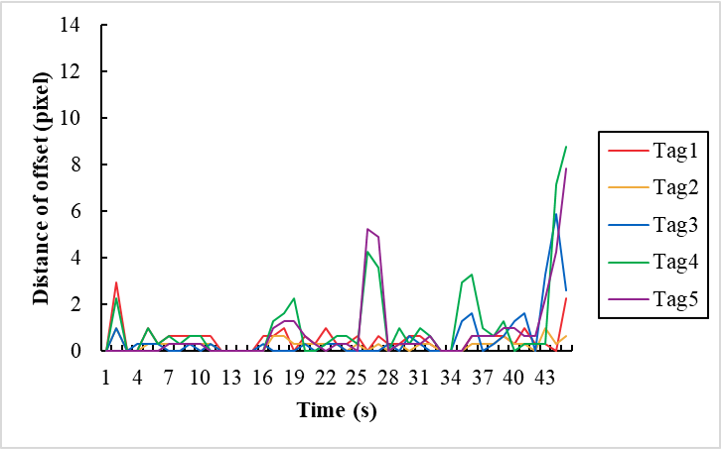}
			\centerline{(b) after}
		\end{minipage}
		\caption{Comparison of clustering effect}
		\label{fig}
	\end{figure}

	Through statistics, we find that the offset distance is significantly improved after improvement. Among them, the deviation situation of label 3 and 4 is significantly optimized, and the fluctuation of 15s almost disappears 13s before the improvement. While the fluctuation in the first 27s of the improvement is also optimized. The fluctuation at 39s and 41s before the improvement also drops a lot. Label 2 has some fluctuation at 27s before the improvement, while the deviation is almost close to 0 after the improvement. Before the improvement of label 1, there is a certain degree of deviation after 31s, and the deviation level off after the improvement.
	
	\section{Conclusion}
	This paper introduces Local Adaptive Clustering method into real-time homography mapping. Adaptive locality is used to reduce the drift of the label in the switching template, and the process of optimizing the matching template is combined with the principle of spatial locality, which reduces the matching time. Experiments show that the method of template segmentation can effectively improve the interior point rate of mapping transformation, and estimate the current frame according to the template of the previous frame can improve the running speed. In summary, locality is able to significantly improve the performance and efficiency issues of matching.
	
	\section*{Acknowledgment}
	This study is partially supported by the National Natural Science Foundation of China (6210071278), the National Key R\&D Program of China (No.2020YFB2103402), the Fundamental Research Funds for the Central Universities (buctrc202123, JD2327). Thank you for the support from HAWKEYE Group.

\end{document}